\documentclass[11pt,a4paper]{article}
\usepackage[hyperref]{acl2018}
\usepackage{times}
\usepackage{latexsym}

\usepackage{url}
\usepackage{graphicx}
\usepackage{amsmath}
\usepackage{multirow}
\graphicspath{{}}
\DeclareGraphicsExtensions{.png}

\aclfinalcopy 
\def\aclpaperid{857} 

\title{Context-Aware Neural Machine Translation Learns Anaphora Resolution}

\author{
  Elena Voita \\
  Yandex, Russia \\
  University of Amsterdam, Netherlands \\
  {\tt lena-voita@yandex-team.ru} \\
  \And
  Pavel Serdyukov \\
  Yandex, Russia \\
  {\tt pavser@yandex-team.ru}
  \AND
  Rico Sennrich \\
  University of Edinburgh, Scotland \\
  University of Zurich, Switzerland \\
  {\tt rico.sennrich@ed.ac.uk} \\
  \And
  Ivan Titov \\
  University of Edinburgh, Scotland \\
  University of Amsterdam, Netherlands \\
  {\tt ititov@inf.ed.ac.uk} \\ \\
}

\date{}

\begin{document}
\maketitle
\begin{abstract}
  Standard machine translation systems
  process sentences in isolation and hence ignore extra-sentential information, even though extended context can both prevent mistakes in ambiguous cases and improve translation coherence. We introduce a context-aware neural machine translation model
  designed in such way that the flow of information from
  the extended context to the translation model can be controlled and analyzed.
  We experiment with an English-Russian subtitles dataset,
  and observe that much of what is captured by our model 
  deals with improving pronoun translation. We measure correspondences between induced attention distributions and coreference relations and observe that the model implicitly captures anaphora. It is consistent with gains for sentences where pronouns need to be gendered in translation. 
  Beside improvements in anaphoric cases, the model also improves in overall BLEU, both over its context-agnostic version (+0.7) and over simple concatenation of the context and source sentences (+0.6).

\end{abstract}

\section{Introduction}

It has long been argued that handling discourse phenomena is important in translation~\cite{10.2307/40006919,hardmeier2012}. Using extended context, beyond the single source sentence, should in principle be beneficial in ambiguous cases and also ensure that generated translations are coherent.
Nevertheless, machine translation systems typically ignore discourse phenomena and translate sentences in isolation.

Earlier research on this topic focused on handling specific phenomena, such as translating pronouns~\cite{lenagard-koehn:2010:WMT,hardmeier2010,hardmeier-EtAl:2015:DiscoMT}, discourse connectives~\cite{AMTA-2012-Meyer}, verb tense~\cite{gong-EtAl:2012:EMNLP-CoNLL}, increasing lexical consistency~\cite{Carpuat:2009:OTP:1621969.1621974,tiedemann10,gong-zhang-zhou:2011:EMNLP}, or topic adaptation~\cite{su-EtAl:2012:ACL2012,hasler-EtAl:2014:EACL}, with special-purpose features engineered to model these phenomena. However, with traditional statistical machine translation being largely supplanted with neural machine translation (NMT) models trained in an end-to-end fashion, an alternative is to directly provide additional context to an NMT system at training time and hope that it will succeed in inducing relevant predictive features~\cite{jean_does_2017,wang_exploiting_2017,tiedemann_neural_2017,DBLP:journals/corr/abs-1711-00513}.  

While the latter approach, using context-aware NMT models, has demonstrated to yield performance improvements, it is still not clear what kinds of discourse phenomena are successfully handled by the NMT systems and, importantly, how they are modeled. Understanding this would inform development of future discourse-aware NMT models, as it will suggest what kind of inductive biases need to be encoded in the architecture or which linguistic features need to be exploited. 

In our work we aim to enhance our understanding of the modelling of selected discourse phenomena in NMT. To this end, we construct a simple discourse-aware model, demonstrate that it achieves improvements over the discourse-agnostic  baseline on an English-Russian subtitles dataset~\cite{LISON18.294}  and study which
context information is being captured in the model. 
Specifically, we start with the Transformer~\cite{attention-is-all-you-need},
a state-of-the-art model for context-agnostic NMT, and modify it in such way
that it can handle additional context.
%
In our model, a source sentence and a context sentence are first encoded independently, and then a single attention layer, in a combination with a gating function, is used to produce a context-aware representation of the source sentence. 
The information from context can only flow through this attention layer. When compared to simply concatenating input sentences, as proposed 
by \newcite{tiedemann_neural_2017}, our architecture appears both more accurate (+0.6 BLEU) and also guarantees that the contextual information cannot bypass the attention layer and hence remain undetected in our analysis.

We analyze what types of contextual information are exploited by the translation model. While studying the attention weights, we observe that much of the information captured by the model has to do with pronoun translation. It is not entirely surprising, as we consider translation from a language without grammatical gender (English) to a language with grammatical gender (Russian). For Russian, translated pronouns need to agree in gender with their antecedents.
Moreover, since in Russian  verbs agree with subjects  in gender and 
adjectives also agree in gender with pronouns in certain frequent  constructions, 
 mistakes in translating pronouns have a major effect on the words in the produced sentences. Consequently, the standard cross-entropy training objective sufficiently rewards the model for improving pronoun translation and extracting relevant information from the context.


We use automatic co-reference systems and human annotation to isolate anaphoric cases. We observe even more substantial improvements in performance 
 on these subsets. By comparing attention distributions induced by our model against co-reference links, we conclude that the model implicitly captures coreference phenomena, even without having any kind of specialized features which could help it in this subtask. These observations also suggest  potential directions for future work. For example, effective co-reference systems
 go beyond relying simply on embeddings of contexts. One option
 would be to integrate `global' features summarizing properties of
 groups of mentions predicted as linked in a document~\cite{wiseman2016global}, or to use latent relations to trace entities across  documents~\cite{ji2017dynamic}.
 Our key contributions can be summarized as follows:
 \begin{itemize}
     \item we introduce a context-aware neural model, which is effective and has a sufficiently simple and interpretable interface between the context and the rest of the translation model;
     \item we analyze the flow of information from the context and identify pronoun translation as the key phenomenon captured by the model;
     \item by comparing to automatically predicted or human-annotated coreference relations, we observe that the model implicitly captures anaphora.

 \end{itemize}


\section{Neural Machine Translation}
Given a source sentence $\textbf{x}=(x_1,\ x_2,\ \dots, x_S)$ and a target sentence $\textbf{y}=(y_1,\ y_2,\ \dots, y_T)$, NMT models predict words in the target sentence, word by word.

Current NMT models mainly have an encoder-decoder structure. The encoder maps an input sequence of symbol representations $\textbf{x}$ to a sequence of distributed representations $\textbf{z}=(z_1,\ z_2,\ \dots, z_S)$. Given $\textbf{z}$, a neural decoder generates the corresponding target sequence of symbols $\textbf{y}$ one element at a time. 

\textbf{Attention-based NMT} The encoder-decoder framework with attention has been proposed by \citet{bahdanau} and has become the de-facto standard in NMT. The model consists of encoder and decoder recurrent networks and an attention mechanism. The attention mechanism selectively focuses on parts of the source sentence during translation, and the attention weights specify the proportions with which information from different positions is combined. 

\textbf{Transformer} \citet{attention-is-all-you-need} proposed an architecture that avoids recurrence completely. The Transformer follows an encoder-decoder architecture using stacked self-attention and 
fully connected layers for both the encoder and decoder. An important advantage of the Transformer is that it is more parallelizable and faster to train than recurrent encoder-decoder models.

From the source tokens, learned embeddings 
are generated and then modified using positional encodings.
The encoded word embeddings are then used as input to the encoder which consists of $N$ layers each containing two sub-layers: (a) a multi-head attention mechanism, and (b) a feed-forward network. 

The self-attention mechanism first computes attention weights: i.e., for each word, it computes a distribution over all words (including itself). This distribution is then used to compute a new representation of that word:  this new representation is set to an expectation (under the  attention distribution specific to the word) of word representations from the layer below. In multi-head attention,
this process is repeated $h$ times with different representations and the result is concatenated. 

The second component of each layer of the Transformer network is a feed-forward network. The authors propose using a two-layered network with the ReLU activations.

Analogously, each layer of the decoder contains the two sub-layers mentioned above as well as an additional multi-head attention sub-layer that receives input from the corresponding encoding layer. 

In the decoder,
the 
attention is masked to prevent future positions from being attended to, or in other words, to prevent illegal leftward information flow. See \citet{attention-is-all-you-need} for additional details. 

The proposed architecture 
reportedly improves over the previous best results on the WMT 2014 English-to-German and English-to-French translation tasks, 
and we verified its strong performance on our data set in preliminary experiments.
Thus, we consider it a strong state-of-the-art baseline for our experiments. Moreover, as 
the Transformer is attractive in practical NMT applications because of 
 its parallelizability and training efficiency, integrating extra-sentential information in Transformer is important from the engineering perspective. As we will see in Section~\ref{sect:exp}, previous techniques developed for recurrent encoder-decoders do not appear effective for the Transformer.

\section{Context-aware model architecture}

\begin{figure}[t!]
\center{\includegraphics[scale=0.28]{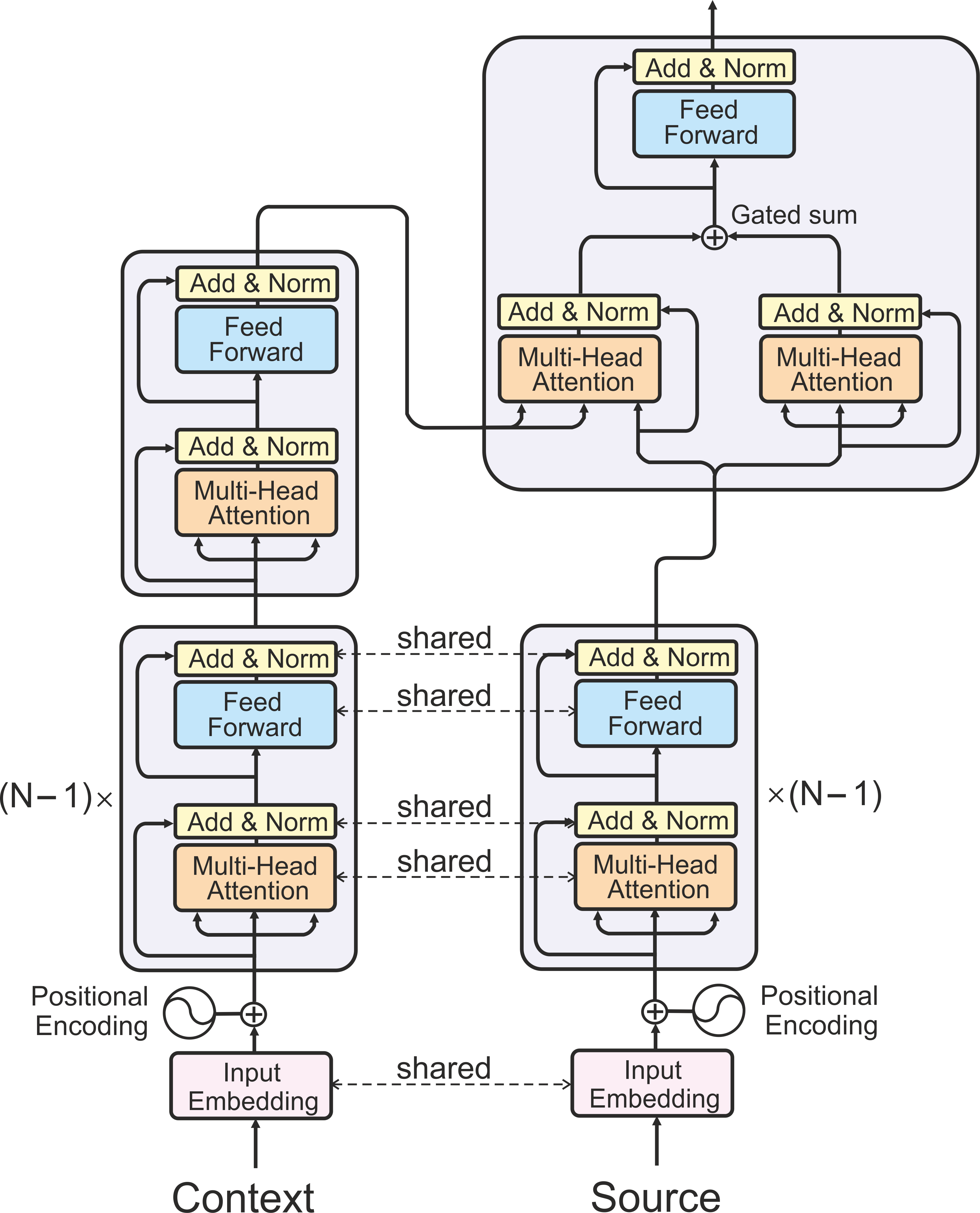}}
\caption{Encoder of the discourse-aware model} 
\label{fig:model_encoder}
\end{figure}

Our model is based on Transformer architecture \citep{attention-is-all-you-need}. We leave Transformer's decoder intact while incorporating context information on the encoder side (Figure~\ref{fig:model_encoder}).

\textbf{Source encoder:} The encoder is composed of a stack of $N$ layers. The first $N-1$ layers are identical and represent the original layers of Transformer's encoder. The last layer incorporates contextual information as shown in Figure~\ref{fig:model_encoder}. In addition to multi-head self-attention it has a block which performs multi-head attention over the output of the context encoder stack. The outputs of the two attention mechanisms are combined via a gated sum.
More precisely, let $c_{i}^{(s-attn)}$ be the output of the multi-head self-attention, $c_i^{(c-attn)}$  the output of the multi-head attention to context, $c_i$~ their gated sum, and $\sigma$ the logistic sigmoid function, then
\begin{equation}
g_i = \sigma\left(W_{g}\left[c_{i}^{(s-attn)}, c_{i}^{(c-attn)}\right] + b_{g}\right)
\end{equation}
\vspace{-1ex}
\begin{equation}
c_i = g_i \odot c_{i}^{(s-attn)} + (1 - g_i) \odot c_{i}^{(c-attn)}
\end{equation}

\textbf{Context encoder:} The context encoder is composed of a stack of $N$ identical layers and replicates the original Transformer encoder. In contrast to related work \cite{jean_does_2017,wang_exploiting_2017},  we found in preliminary experiments that using
separate encoders does not yield an accurate model.
Instead we share the parameters of the first $N-1$ layers with the source encoder.

Since major proportion of the context encoder's parameters are shared with the source encoder, we add a special token (let us denote it {\verb|<bos>|}) to the beginning of context sentences, but not source sentences, to let the shared layers know whether it is encoding a source or a context sentence.

\section{Experiments}
\label{sect:exp}
\subsection{Data and setting}
We use the publicly available OpenSubtitles2018 corpus~\cite{LISON18.294} for English and Russian.\footnote{\url{http://opus.nlpl.eu/OpenSubtitles2018.php}} 
As described in the appendix, we apply data cleaning and randomly choose 2 million training instances from the resulting data. For development and testing, we randomly select two subsets of 10000 instances from movies not encountered in training.\footnote{The resulting data sets are freely available at \url{http://data.statmt.org/acl18_contextnmt_data/}}
Sentences were encoded using byte-pair encoding~\cite{sennrich-bpe}, with source and target vocabularies of about 32000 tokens.
We generally used the same parameters and optimizer as in the original Transformer~\cite{attention-is-all-you-need}. 
The hyperparameters, preprocessing and training details are provided in the supplementary material.







\section{Results and analysis}


We start by experiments motivating the setting and verifying that the improvements are indeed genuine, i.e. they come from inducing predictive features of the context. In subsequent section \ref{sect:analysis}, we analyze the features induced by the context encoder and perform error analysis. 








\subsection{Overall performance}

We use the traditional automatic metric BLEU on a general test set to get an estimate of the overall performance of the discourse-aware model, before turning to more targeted evaluation in the next section.
%
%
%
We provide results in Table~\ref{tab:common_bleu}.\footnote{We use bootstrap resampling \cite{riezler-maxwell:2005:ACL2005} for significance testing} The `baseline' is the discourse-agnostic version of the Transformer.  As another baseline we use the standard Transformer applied to the concatenation of the previous and source sentences, as proposed by
\newcite{tiedemann_neural_2017}. \newcite{tiedemann_neural_2017} only used a special symbol to mark where the context sentence ends and the source sentence begins. This technique performed badly with the non-recurrent Transformer architecture in preliminary experiments, resulting in a substantial degradation of performance (over 1 BLEU). Instead, we use a binary flag at every word position in our concatenation baseline telling the encoder whether the word belongs to the context sentence or to the source sentence.


We consider two versions of our discourse-aware model: one using the previous sentence as the context, another one relying on the next sentence. We hypothesize that both the previous and the next sentence provide a similar amount of additional clues about the topic of the text, whereas for discourse phenomena such as anaphora, discourse relations and elliptical structures, the previous sentence is more important.

First, we observe that our best model is the one using a context encoder for the previous sentence: it achieves  0.7 BLEU
improvement over the discourse-agnostic model. We also notice that, unlike  the previous sentence, the next sentence does not appear beneficial. This is a first indicator that discourse phenomena are the main reason for the observed improvement, rather than topic effects.
Consequently, we focus solely on using  the previous sentence in all subsequent experiments. 

Second, we observe that the concatenation baseline appears less accurate than the introduced context-aware model. This result suggests that our model is not only more amendable to analysis but also potentially more effective than using concatenation.


\begin{table}[t!]
\begin{center}
\begin{tabular}{|l|c|}
\hline \bf \bf model  & \bf BLEU \\ \hline
baseline & 29.46  \\ 
concatenation (previous sentence) & 29.53 \\
context encoder (previous sentence) & \bf 30.14 \\
context encoder (next sentence) & 29.31 \\
context encoder (random context) & 29.69 \\
\hline
\end{tabular}
\end{center}
\vspace{-1ex}
\caption{\label{tab:common_bleu} Automatic evaluation: BLEU. Significant differences at $p<0{.}01$ are in bold. }
\vspace{-1ex}
\end{table}

In order to verify that our improvements are genuine, we also evaluate our model (trained with the previous sentence as context) on the same test set with shuffled context sentences. It can be seen that the performance drops significantly when a real context sentence is replaced with a random one. This confirms that the model does rely on context information to achieve the improvement in translation quality, and is not merely better regularized. However, the model is 
robust towards being shown a random context and obtains a performance similar to the context-agnostic baseline.


\subsection{Analysis}
\label{sect:analysis}

In this section we investigate what types of contextual information are exploited by the  model.
We study the distribution of attention to context and perform analysis on specific subsets of the test data. 
Specifically the research questions we seek to answer are as follows:
%
%
%
\begin{itemize}
\item For the translation of which words does the model rely on contextual history most?
\vspace{-1ex}
\item Are there any non-lexical patterns affecting attention to context, such as sentence length and word position?
\vspace{-1ex}
\item Can the context-aware NMT system  implicitly learn coreference phenomena without any feature engineering?
\end{itemize}

Since all the attentions in our model are multi-head, by \textit{attention weights} we refer to an average over heads of per-head attention weights.

First, we would like to identify a \textit{useful} attention mass coming to context. 
We analyze the attention maps between source and context, and find that the model mostly attends to {\verb|<bos>|} and {\verb|<eos>|} context tokens, and much less often attends to words. Our hypothesis is that the model has found a way to take no information from context by looking at uninformative tokens, and it attends to words only when it wants to pass some contextual information to the source sentence encoder. Thus we define \textit{useful} contextual attention mass as sum of attention weights to context words, excluding {\verb|<bos>|} and {\verb|<eos>|} tokens and punctuation.

\subsubsection{Top words depending on context}
We analyze the distribution of attention to context for individual source words to see for which words the model depends most on contextual history. We compute the overall average attention to context words for each source word in our test set. We do the same for source words at positions higher than first. We filter out words that occurred less than 10 times in a test set. The top 10 words with the highest average attention to context words are provided in Table~\ref{tab:top_avg_attn}.

\begin{table}[t!]
\begin{center}
\begin{tabular}{|ccc|ccc|}
\hline \bf word  & \bf attn & \bf pos & \bf word & \bf attn & \bf pos \\ \hline

it & 0.376 & 5.5 & it & 0.342 & 6.8\\
yours & 0.338 & 8.4 & yours & 0.341 & 8.3 \\
yes & 0.332 & 2.5 & ones & 0.318 & 7.5\\
i & 0.328 & 3.3 & 'm & 0.301 & 4.8\\
yeah & 0.314 & 1.4 & you & 0.287 & 5.6\\
you & 0.311 & 4.8 & am & 0.274 & 4.4\\
ones & 0.309 & 8.3 & i & 0.262 & 5.2\\
'm & 0.298 & 5.1 & 's & 0.260 & 5.6\\
wait & 0.281 & 3.8 & one & 0.259 & 6.5\\
well & 0.273 & 2.1 & won & 0.258 & 4.6\\
\hline
\end{tabular}
\end{center}
\vspace{-1ex}
\caption{\label{tab:top_avg_attn} Top-10 words with the highest average attention to context words. \textit{attn} gives an average attention to context words, \textit{pos} gives an average position of the source word. Left part is for words on all positions, right~--- for words on positions higher than first.}
\vspace{-2ex}
\end{table}

An interesting finding is that contextual attention is high for the translation of ``it'', ``yours'', ``ones'', ``you'' and ``I'', which are indeed very ambiguous out-of-context when translating into Russian. For example, ``it'' will be translated as third person singular masculine, feminine or neuter, or third person plural depending on its antecedent. ``You'' can be second person singular impolite or polite, or plural. Also, verbs must agree in gender and number with the translation of ``you''.

It might be not obvious why ``I'' has high contextual attention, as it is not ambiguous itself. However, in past tense, verbs must agree with ``I'' in gender, so to translate  past tense sentences properly, the source encoder must predict speaker gender, and the context may provide useful indicators.

Most surprising is the appearance of ``yes'', ``yeah'', and ``well'' in the list of context-dependent words, similar to the finding by \citet{tiedemann_neural_2017}. 
We note that these words mostly appear in sentence-initial position, and in relatively short sentences. If only words after the first are considered, they disappear from the top-10 list. We hypothesize that the amount of attention to context not only depends on the words themselves, but also on factors such as sentence length and position, and we test this hypothesis in the next section.

\subsubsection{Dependence on sentence length and position}
\begin{figure}[t!]
\center{\includegraphics[scale=0.35]{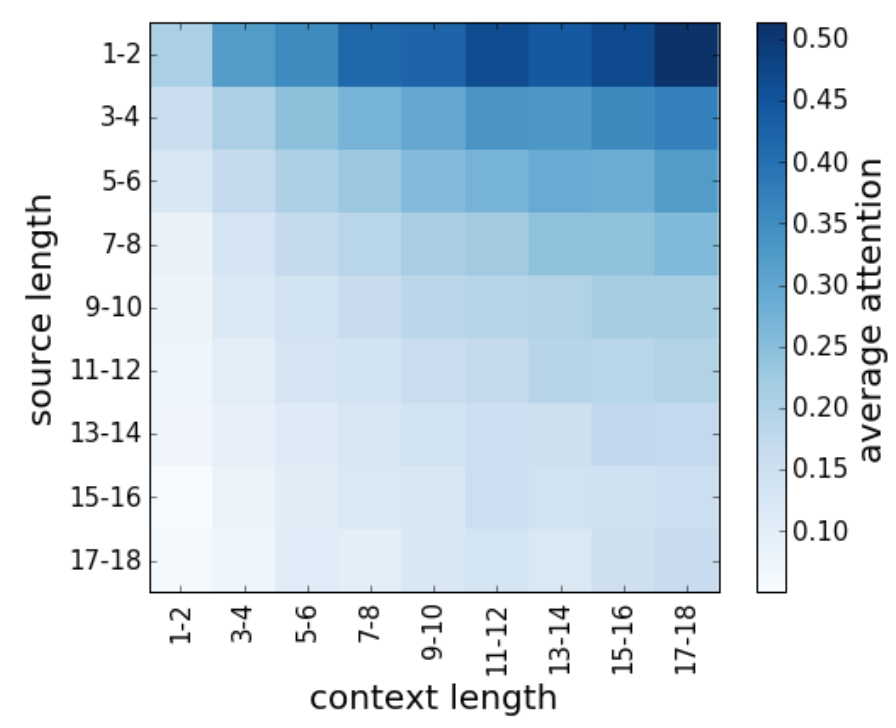}}
\vspace{-1ex}
\caption{Average attention to context words vs. both source and context length}
\vspace{-1ex}
\label{fig:attn_from_bothlen}
\end{figure}

We compute useful attention mass coming to context by averaging  over source words. Figure~\ref{fig:attn_from_bothlen} illustrates the dependence of this average attention mass on sentence length. We observe a disproportionally high attention on context for short sentences, and a positive correlation between the average contextual attention and context length.




\begin{figure}[t!]
\center{\includegraphics[scale=0.33]{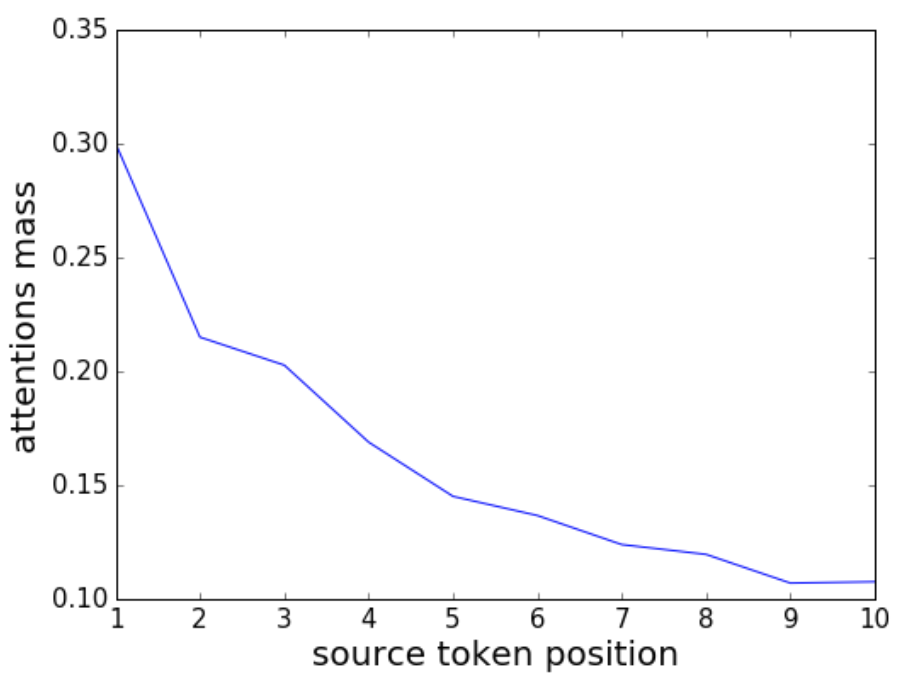}}
\vspace{-1ex}
\caption{Average attention to context  vs. source token position}
\vspace{-1ex}
\label{fig:attn_from_src_tok_position}
\end{figure}

It is also interesting to see the importance given to the context at different positions in the source sentence. We compute an average attention mass to context for a set of 1500 sentences of the same length. As can be seen in Figure~\ref{fig:attn_from_src_tok_position}, words at the beginning of a source sentence tend to attend to context more than words at the end of a sentence. This correlates with standard view that English sentences present hearer-old material before hearer-new.

\begin{figure}[t!]
\center{\includegraphics[scale=0.33]{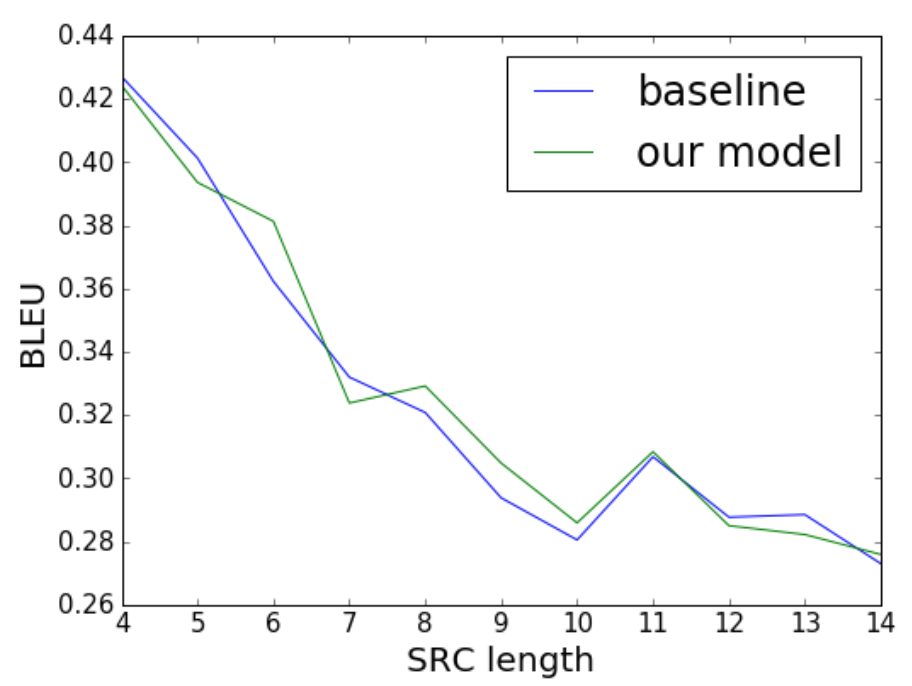}}
\vspace{-1ex}
\caption{BLEU score vs. source sentence length}
\vspace{-1ex}
\label{fig:bleu_from_src_length}
\end{figure}

There is a clear (negative) correlation between sentence length and the amount of attention placed on contextual history, and between token position and the amount of attention to context, which suggests that context is especially helpful at the beginning of a sentence, and for shorter sentences.
However, Figure~\ref{fig:bleu_from_src_length} shows that there is no straightforward dependence of BLEU improvement on source length. This means that while attention on context is disproportionally high for short sentences, context does not seem disproportionally more useful for these sentences.

\subsection{Analysis of pronoun translation}
The analysis of the attention model indicates that the model attends heavily to the contextual history for the translation of some pronouns. Here, we investigate whether this context-aware modelling results in empirical improvements in translation quality, and whether the model learns structures related to anaphora resolution.

\begin{table*}
\centering
\begin{tabular}{|l|cc|cc|c|}
\hline \bf pronoun  & \textbf{N} & \textbf{\#pronominal antecedent} & \bf baseline & \bf our model & \textbf{difference} \\ \hline
it & 11128 & 6604 & 25.4 & 26.6 & \textbf{+1.2}  \\
you & \phantom{0}6398 & 5795 & 29.7 & 30.8 & \textbf{+1.1} \\
yours & \phantom{0}2181 & 2092 & 24.1 & 25.2 & \textbf{+1.1} \\
I & \phantom{0}8205 & 7496 & 30.1 & 30.0 & -0.1 \\
\hline
\end{tabular}
\vspace{-1ex}
\caption{\label{tab:any_coreference_bleu} BLEU for test sets with coreference between pronoun and a word in context sentence. We show both \textit{N}, the total number of instances in a particular test set, and number of instances with \textit{pronominal antecedent}. Significant BLEU differences are  in bold.}
\end{table*}

\subsubsection{Ambiguous pronouns and translation quality}

Ambiguous pronouns are relatively sparse in a general-purpose test set, and previous work has designed targeted evaluation of pronoun translation \cite{hardmeier-EtAl:2015:DiscoMT,miculicichwerlen-popescubelis:2017:DiscoMT,DBLP:journals/corr/abs-1711-00513}. However, we note that in Russian, grammatical gender is not only marked on pronouns, but also on adjectives and verbs. Rather than using a pronoun-specific evaluation, we present results with BLEU on test sets where we hypothesize context to be relevant, specifically sentences containing co-referential pronouns.
We feed Stanford CoreNLP open-source coreference resolution system \cite{manning2014stanford} with pairs of sentences to find examples where there is a link between one of the pronouns under consideration and the context.
We focus on anaphoric instances of ``it'' (this excludes, among others, pleonastic uses of "it"), and instances of the pronouns ``I'', ``you'', and ``yours'' that are coreferent with an expression in the previous sentence. All these pronouns express ambiguity in the translation into Russian, and the model has learned to attend to context for their translation (Table~~\ref{tab:top_avg_attn}).
To combat data sparsity, the test sets are extracted from large amounts of held-out data of OpenSubtitles2018.
Table~\ref{tab:any_coreference_bleu} shows BLEU scores for the resulting subsets.

First of all, we see that most of the antecedents in these test sets are also pronouns. Antecedent pronouns should not be particularly informative for translating the source pronoun. Nevertheless, even with such contexts, improvements are generally larger than on the overall test set. 

\begin{table}[t!]
\begin{center}
\begin{tabular}{|l |c|cc|c|}
\hline \bf word   & \textbf{N} & \bf baseline & \bf our model & \bf diff. \\ \hline
it & 4524 & 23.9 & 26.1 & \textbf{+2.2}  \\
you & \phantom{0}693 & 29.9 & 31.7 & \textbf{+1.8} \\
I & \phantom{0}709 & 29.1 & 29.7 & \textbf{+0.6} \\
\hline
\end{tabular}
\end{center}
\vspace{-1ex}
\caption{\label{tab:noun_coreference_bleu} BLEU for test sets of pronouns having a nominal antecedent in context sentence. \textit{N}: number of examples in the test set.}
\vspace{-1ex}
\end{table}

\begin{table}[t!]
\begin{center}
\begin{tabular}{|l|c|cc|c|}
\hline \bf type  & \textbf{N} & \bf baseline & \bf our model & \bf diff. \\ \hline
masc. & 2509 & 26.9 & 27.2 & \textbf{+0.3} \\
fem. & 2403 & 21.8 & 26.6 & \textbf{+4.8} \\
neuter & \phantom{0}862 & 22.1 & 24.0 & \textbf{+1.9} \\
plural & 1141 & 18.2 & 22.5 & \textbf{+4.3} \\
\hline
\end{tabular}
\end{center}
\vspace{-1ex}
\caption{\label{tab:it_bleu} BLEU for test sets of pronoun ``it'' having a nominal antecedent in context sentence. \textit{N}: number of examples in the test set.}
\vspace{-1ex}
\end{table}

When we focus on sentences where the antecedent for pronoun under consideration contains a noun, we observe  even larger improvements (Table~\ref{tab:noun_coreference_bleu}).
  Improvement is smaller for ``I'', but we note that verbs with first person singular subjects mark gender only in the past tense, which limits the impact of correctly predicting gender. In contrast, different types of ``you'' (polite/impolite, singular/plural) lead to different translations of the pronoun itself plus related verbs and adjectives, leading to a larger jump in performance. 
  Examples of nouns co-referent with ``I'' and ``you'' include names, titles (``Mr.'', ``Mrs.'', ``officer''), terms denoting family relationships (``Mom'', ``Dad''), and terms of endearment (``honey'', ``sweetie'').
  Such nouns can serve to disambiguate number and gender of the speaker or addressee, and mark the level of familiarity between them.
  

The most interesting case is translation of ``it'', as ``it'' can have many different translations into Russian, depending on the grammatical gender of the antecedent.
In order to disentangle these cases,
 we train the Berkeley aligner on 10m sentences and use the trained model to divide the test set with ``it'' referring to a noun into test sets specific to each gender and number. Results are in Table~\ref{tab:it_bleu}.

We see an improvement of 4-5 BLEU for sentences where ``it'' is translated into a feminine or plural pronoun by the reference. For cases where ``it'' is translated into a masculine pronoun, the improvement is smaller because the masculine gender is more frequent, and the context-agnostic baseline tends to translate the pronoun ``it'' as masculine. 


\subsubsection{Latent anaphora resolution} 

The results in Tables~\ref{tab:noun_coreference_bleu} and~\ref{tab:it_bleu} suggest that the context-aware model exploits information about the antecedent of an ambiguous pronoun.
We hypothesize that we can interpret the model's attention mechanism as a latent anaphora resolution, and perform experiments to test this hypothesis.

For test sets from Table~\ref{tab:noun_coreference_bleu}, we find an antecedent noun phrase (usually a determiner or a possessive pronoun followed by a noun) using Stanford CoreNLP \cite{manning-EtAl:2014:P14-5}. We select only examples where a noun phrase contains a single noun to simplify our analysis. Then we identify which token receives the highest attention weight (excluding {\verb|<bos>|} and {\verb|<eos>|} tokens and punctuation).
If this token falls within the antecedent span, then we treat it as agreement (see Table~\ref{tab:it_coreference_agreement}).

\begin{table}[t!]
\begin{center}
\begin{tabular}{|l|c|c|c|c|}
\hline
\multirow{2}{*}{\bf pronoun}  & \multicolumn{4}{c|}{\bf agreement (in \%)}\\
& \bf random & \bf first & \bf last & \bf attention \\
\hline
it & 69 & 66 & 72 & 69 \\
you & 76 & 85 & 71 & 80 \\
I & 74 & 81 & 73 & 78 \\
\hline
\end{tabular}
\end{center}
\vspace{-1ex}
\caption{\label{tab:it_coreference_agreement} Agreement with CoreNLP for test
 sets of pronouns having a nominal antecedent in context sentence (\%).}
 \vspace{-1ex}
\end{table}

One natural question might be: does the attention component in our model genuinely learn to perform anaphora resolution, or does it capture some simple heuristic (e.g., pointing to the last noun)? To answer this question, we consider several baselines: choosing  a random, last or first noun from the context sentence as an antecedent.


Note that an agreement of the last noun for ``it'' or the first noun for ``you'' and ``I'' is very high. This is partially due to the fact that most  context sentences have only one noun. For these examples a random and last predictions are always correct, meanwhile attention does not always pick a noun as the most relevant word in the context. To get a more clear picture let us now concentrate only on examples where there is more than one noun in the context (Table~\ref{tab:it_coreference_agreement_mult}). 
We can now see that the attention weights are in much better agreement with the coreference system than any of the heuristics. This indicates that the model is indeed performing anaphora resolution.

\begin{table}[t!]
\begin{center}
\begin{tabular}{|l|c|c|c|c|}
\hline 
\multirow{2}{*}{\bf pronoun}  & \multicolumn{4}{c|}{\bf agreement (in \%)}\\
& \bf random & \bf first & \bf last & \bf attention \\
\hline
it & 40 & 36 & 52 & 58 \\
you & 42 & 63 & 29 & 67 \\
I & 39 & 56 & 35 & 62 \\
\hline
\end{tabular}
\end{center}
\vspace{-1ex}
\caption{\label{tab:it_coreference_agreement_mult} Agreement with CoreNLP for test sets of pronouns having a nominal antecedent in context sentence (\%). Examples with $\geq$1 noun in context sentence.}
\vspace{-1ex}
\end{table}

While agreement with CoreNLP is encouraging, we are aware that coreference resolution by CoreNLP is imperfect and partial agreement with it may not necessarily indicate that the attention is particularly accurate. In order to control for this, we asked human annotators to manually evaluate 500 examples from the test sets where CoreNLP predicted that ``it'' refers to a noun in the context sentence. More precisely, we picked random 500 examples from the test set with ``it'' from Table~\ref{tab:it_coreference_agreement_mult}. We marked the pronoun in a source which CoreNLP found anaphoric. Assessors were given the source and context sentences and were asked to mark an antecedent noun phrase for a marked pronoun in a source sentence or say that there is no antecedent at all. We then picked those examples where assessors found a link from ``it'' to some noun in context ($79\%$ of all examples). Then we evaluated  agreement of CoreNLP and our model with the ground truth links. We also report the performance of the best heuristic for ``it'' from our previous analysis (i.e. last noun in context). The results are provided in Table~\ref{tab:it_coreference_mult_human}.


\begin{table}[t!]
\begin{center}
\begin{tabular}{|l|c|}
\hline  & \bf agreement (in \%) \\ \hline
CoreNLP & 77 \\
attention & 72 \\
last noun & 54 \\
\hline
\end{tabular}
\end{center}
\vspace{-1ex}
\caption{\label{tab:it_coreference_mult_human} Performance of CoreNLP and our model's attention mechanism compared to human assessment. Examples with $\geq$1 noun in context sentence.}
\vspace{-1ex}
\end{table}

\begin{figure}[t!]
\center{\includegraphics[scale=0.4]{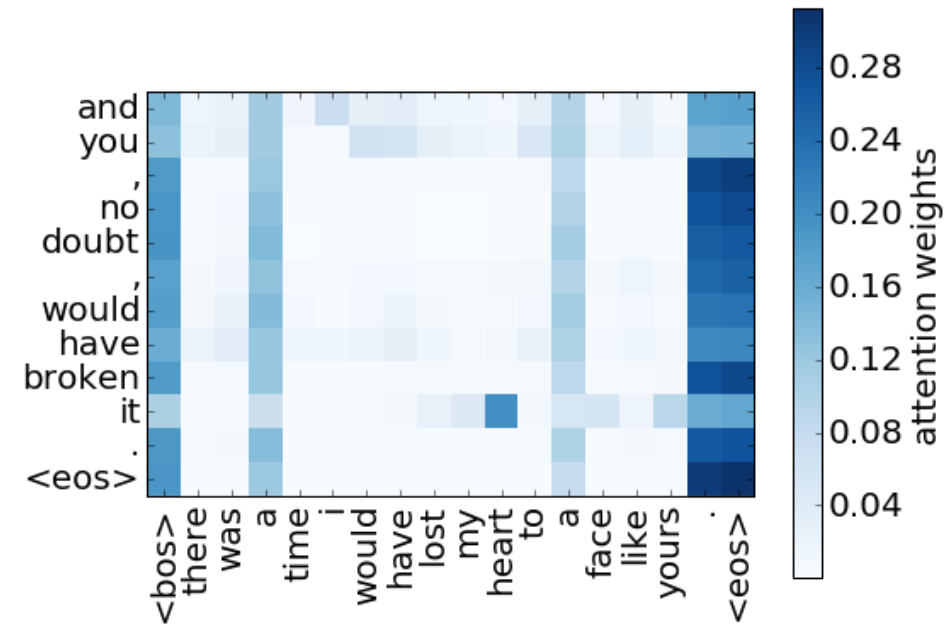}}
\vspace{-1ex}
\caption{An example of an attention map between source and context. On the $y$-axis are the source tokens, on the $x$-axis the context tokens. Note the high attention between ``it'' and its antecedent ``heart''.}
\vspace{-1ex}
\label{fig:it_coreference_example}
\end{figure}

The agreement between our model and the ground truth  is 72$\%$. Though 5\% below the coreference system, this is a lot higher than the best heuristic (+18\%). This confirms our conclusion that our model performs latent anaphora resolution. 
Interestingly, the patterns of mistakes are quite different for CoreNLP and our model (Table~\ref{tab:it_coreference_mult_human_compare}). We also present one example (Figure~\ref{fig:it_coreference_example}) where the attention correctly predicts anaphora while CoreNLP fails.
 Nevertheless, there is  room for improvement, and improving the attention component is likely to boost  translation performance.  

\begin{table}[t!]
\begin{center}
\begin{tabular}{|l|cc|}
\hline
& \multicolumn{2}{c|}{CoreNLP}\\
&  right & wrong \\
\hline
attn right & 53 & 19 \\
attn wrong & 24 & \phantom{0}4 \\
\hline
\end{tabular}
\end{center}
\vspace{-1ex}
\vspace{-1ex}
\caption{\label{tab:it_coreference_mult_human_compare}
Performance of CoreNLP and our model's attention mechanism compared to human assessment (\%). Examples with $\geq$1 noun in context sentence.}
\vspace{-1ex}
\end{table}


\section{Related work}
Our analysis focuses on how our context-aware neural model implicitly captures anaphora.
Early work on anaphora phenomena in statistical machine translation has relied on external systems for coreference resolution \cite{lenagard-koehn:2010:WMT,hardmeier2010}. Results were mixed, and the low performance of coreference resolution systems was identified as a problem for this type of system. Later work by \citet{hardmeier-tiedemann-nivre:2013:EMNLP} has shown that cross-lingual pronoun prediction systems can implicitly learn to resolve coreference, but this work still relied on external feature extraction to identify anaphora candidates. Our experiments show that a context-aware neural machine translation system can implicitly learn coreference phenomena without any feature engineering.

\citet{tiedemann_neural_2017} and \citet{DBLP:journals/corr/abs-1711-00513} analyze the attention weights of context-aware NMT models. \citet{tiedemann_neural_2017} find some evidence for above-average attention on contextual history for the translation of pronouns, and our analysis goes further in that we are the first to demonstrate that our context-aware model learns latent anaphora resolution through the attention mechanism. This is contrary to \citet{DBLP:journals/corr/abs-1711-00513}, who do not observe increased attention between a pronoun and its antecedent in their recurrent model. We deem our model more suitable for analysis, since it has no recurrent connections and fully relies on the attention mechanism within a single attention layer.

\section{Conclusions}

We introduced a context-aware NMT system which is based on the Transformer architecture. When evaluated on an En-Ru parallel corpus, it outperforms both the context-agnostic baselines and a simple context-aware baseline. We observe that improvements are especially prominent for sentences containing ambiguous pronouns. We also show that the model induces anaphora relations. We believe that further improvements in handling anaphora, and by proxy translation, can be achieved by  incorporating specialized features in the attention model.
Our analysis has focused on the effect of context information on pronoun translation.
Future work could also investigate whether context-aware NMT systems learn other discourse phenomena,
for example whether they improve the translation of elliptical constructions, and markers of discourse relations and information structure.

\section*{Acknowledgments}

We would like to thank Bonnie Webber for helpful discussions and annonymous reviewers for their comments.
The authors also thank David Talbot and Yandex Machine Translation team for helpful discussions and inspiration.
Ivan Titov acknowledges support of the European Research Council
(ERC StG BroadSem 678254) and the Dutch National Science Foundation (NWO
VIDI 639.022.518).
Rico  Sennrich  has  received  funding  from  the Swiss National Science Foundation (105212\_169888).

\bibliography{acl2018}

\begin{thebibliography}{}
\expandafter\ifx\csname natexlab\endcsname\relax\def\natexlab#1{#1}\fi

\bibitem[{Bahdanau et~al.(2015)Bahdanau, Cho, and Bengio}]{bahdanau}
Dzmitry Bahdanau, Kyunghyun Cho, and Yoshua Bengio. 2015.
\newblock Neural machine translation by jointly learning to align and
  translate.
\newblock In {\em Proceedings of the Third International Conference on Learning
  Representations (ICLR 2015)\/}. San Diego.

\bibitem[{{Bawden} et~al.(2018){Bawden}, {Sennrich}, {Birch}, and
  {Haddow}}]{DBLP:journals/corr/abs-1711-00513}
Rachel {Bawden}, Rico {Sennrich}, Alexandra {Birch}, and Barry {Haddow}. 2018.
\newblock {Evaluating Discourse Phenomena in Neural Machine Translation}.
\newblock In {\em {Proceedings of the 16th Annual Conference of the North
  American Chapter of the Association for Computational Linguistics: Human
  Language Technologies}\/}. New Orleans, USA.

\bibitem[{Carpuat(2009)}]{Carpuat:2009:OTP:1621969.1621974}
Marine Carpuat. 2009.
\newblock \href{http://www.aclweb.org/anthology/W09-2404}{{One Translation Per
  Discourse}}.
\newblock In {\em {Proceedings of the Workshop on Semantic Evaluations: Recent
  Achievements and Future Directions}\/}. Association for Computational
  Linguistics, Boulder, Colorado, pages 19--27.
\newblock
  \href{http://www.aclweb.org/anthology/W09-2404}{http://www.aclweb.org/anthology/W09-2404}.

\bibitem[{Gong et~al.(2012)Gong, Zhang, Tan, and
  Zhou}]{gong-EtAl:2012:EMNLP-CoNLL}
Zhengxian Gong, Min Zhang, Chew~Lim Tan, and Guodong Zhou. 2012.
\newblock \href{http://www.aclweb.org/anthology/D12-1026}{N-gram-based tense
  models for statistical machine translation}.
\newblock In {\em Proceedings of the 2012 Joint Conference on Empirical Methods
  in Natural Language Processing and Computational Natural Language
  Learning\/}. Association for Computational Linguistics, Jeju Island, Korea,
  pages 276--285.
\newblock
  \href{http://www.aclweb.org/anthology/D12-1026}{http://www.aclweb.org/anthology/D12-1026}.

\bibitem[{Gong et~al.(2011)Gong, Zhang, and Zhou}]{gong-zhang-zhou:2011:EMNLP}
Zhengxian Gong, Min Zhang, and Guodong Zhou. 2011.
\newblock \href{http://www.aclweb.org/anthology/D11-1084}{{Cache-based
  Document-level Statistical Machine Translation}}.
\newblock In {\em {Proceedings of the 2011 Conference on Empirical Methods in
  Natural Language Processing}\/}. Association for Computational Linguistics,
  Edinburgh, Scotland, UK., pages 909--919.
\newblock
  \href{http://www.aclweb.org/anthology/D11-1084}{http://www.aclweb.org/anthology/D11-1084}.

\bibitem[{Hardmeier(2012)}]{hardmeier2012}
Christian Hardmeier. 2012.
\newblock {Discourse in statistical machine translation: A survey and a case
  study}.
\newblock {\em Discours\/} 11.

\bibitem[{Hardmeier and Federico(2010)}]{hardmeier2010}
Christian Hardmeier and Marcello Federico. 2010.
\newblock {Modelling Pronominal Anaphora in Statistical Machine Translation}.
\newblock In {\em {Proceedings of the seventh International Workshop on Spoken
  Language Translation (IWSLT)}\/}. pages 283--289.

\bibitem[{Hardmeier et~al.(2015)Hardmeier, Nakov, Stymne, Tiedemann, Versley,
  and Cettolo}]{hardmeier-EtAl:2015:DiscoMT}
Christian Hardmeier, Preslav Nakov, Sara Stymne, J{\"o}rg Tiedemann, Yannick
  Versley, and Mauro Cettolo. 2015.
\newblock \href{https://doi.org/10.18653/v1/W15-2501}{{Pronoun-Focused MT and
  Cross-Lingual Pronoun Prediction: Findings of the 2015 DiscoMT Shared Task on
  Pronoun Translation}}.
\newblock In {\em {Proceedings of the Second Workshop on Discourse in Machine
  Translation}\/}. Association for Computational Linguistics, Lisbon, Portugal,
  pages 1--16.
\newblock
  \href{https://doi.org/10.18653/v1/W15-2501}{https://doi.org/10.18653/v1/W15-2501}.

\bibitem[{Hardmeier et~al.(2013)Hardmeier, Tiedemann, and
  Nivre}]{hardmeier-tiedemann-nivre:2013:EMNLP}
Christian Hardmeier, J\"{o}rg Tiedemann, and Joakim Nivre. 2013.
\newblock \href{http://www.aclweb.org/anthology/D13-1037}{Latent anaphora
  resolution for cross-lingual pronoun prediction}.
\newblock In {\em Proceedings of the 2013 Conference on Empirical Methods in
  Natural Language Processing\/}. Association for Computational Linguistics,
  Seattle, Washington, USA, pages 380--391.
\newblock
  \href{http://www.aclweb.org/anthology/D13-1037}{http://www.aclweb.org/anthology/D13-1037}.

\bibitem[{Hasler et~al.(2014)Hasler, Blunsom, Koehn, and
  Haddow}]{hasler-EtAl:2014:EACL}
Eva Hasler, Phil Blunsom, Philipp Koehn, and Barry Haddow. 2014.
\newblock \href{https://doi.org/10.3115/v1/E14-1035}{Dynamic topic adaptation
  for phrase-based mt}.
\newblock In {\em Proceedings of the 14th Conference of the European Chapter of
  the Association for Computational Linguistics\/}. Association for
  Computational Linguistics, Gothenburg, Sweden, pages 328--337.
\newblock
  \href{https://doi.org/10.3115/v1/E14-1035}{https://doi.org/10.3115/v1/E14-1035}.

\bibitem[{Jean et~al.(2017)Jean, Lauly, Firat, and Cho}]{jean_does_2017}
Sebastien Jean, Stanislas Lauly, Orhan Firat, and Kyunghyun Cho. 2017.
\newblock {Does {Neural} {Machine} {Translation} {Benefit} from {Larger}
  {Context}?}
\newblock In {\em {{arXiv}:1704.05135}\/}.
\newblock ArXiv: 1704.05135.

\bibitem[{Ji et~al.(2017)Ji, Tan, Martschat, Choi, and Smith}]{ji2017dynamic}
Yangfeng Ji, Chenhao Tan, Sebastian Martschat, Yejin Choi, and Noah~A Smith.
  2017.
\newblock \href{https://doi.org/10.18653/v1/D17-1195}{Dynamic entity
  representations in neural language models}.
\newblock In {\em Proceedings of the 2017 Conference on Empirical Methods in
  Natural Language Processing\/}. Association for Computational Linguistics,
  Copenhagen, Denmark, pages 1830--1839.
\newblock
  \href{https://doi.org/10.18653/v1/D17-1195}{https://doi.org/10.18653/v1/D17-1195}.

\bibitem[{Kingma and Ba(2015)}]{adam-optimizer}
Diederik Kingma and Jimmy Ba. 2015.
\newblock {Adam: A method for stochastic optimization}.
\newblock In {\em {Proceedings of the {International} {Conference} on
  {Learning} {Representation} (ICLR 2015)}\/}.

\bibitem[{Le~Nagard and Koehn(2010)}]{lenagard-koehn:2010:WMT}
Ronan Le~Nagard and Philipp Koehn. 2010.
\newblock \href{http://www.aclweb.org/anthology/W10-1737}{Aiding pronoun
  translation with co-reference resolution}.
\newblock In {\em Proceedings of the Joint Fifth Workshop on Statistical
  Machine Translation and MetricsMATR\/}. Association for Computational
  Linguistics, Uppsala, Sweden, pages 252--261.
\newblock
  \href{http://www.aclweb.org/anthology/W10-1737}{http://www.aclweb.org/anthology/W10-1737}.

\bibitem[{Lison and Tiedemann(2016)}]{open_subtitles_2016}
Pierre Lison and J{\"o}rg Tiedemann. 2016.
\newblock {OpenSubtitles2016: Extracting Large Parallel Corpora from Movie and
  TV Subtitles}.
\newblock In {\em {Proceedings of the 10th {International} {Conference} on
  {Language} {Resources} and {Evaluation} (LREC 2016)}\/}.

\bibitem[{Lison et~al.(2018)Lison, Tiedemann, and Kouylekov}]{LISON18.294}
Pierre Lison, J\"{o}rg Tiedemann, and Milen Kouylekov. 2018.
\newblock Opensubtitles2018: Statistical rescoring of sentence alignments in
  large, noisy parallel corpora.
\newblock In {\em Proceedings of the Eleventh International Conference on
  Language Resources and Evaluation (LREC 2018)\/}. Miyazaki, Japan.

\bibitem[{Manning et~al.(2014{\natexlab{a}})Manning, Surdeanu, Bauer, Finkel,
  Bethard, and McClosky}]{manning2014stanford}
Christopher Manning, Mihai Surdeanu, John Bauer, Jenny Finkel, Steven Bethard,
  and David McClosky. 2014{\natexlab{a}}.
\newblock \href{https://doi.org/10.3115/v1/P14-5010}{The stanford corenlp
  natural language processing toolkit}.
\newblock In {\em Proceedings of 52nd Annual Meeting of the Association for
  Computational Linguistics: System Demonstrations\/}. Association for
  Computational Linguistics, Baltimore, Maryland, pages 55--60.
\newblock
  \href{https://doi.org/10.3115/v1/P14-5010}{https://doi.org/10.3115/v1/P14-5010}.

\bibitem[{Manning et~al.(2014{\natexlab{b}})Manning, Surdeanu, Bauer, Finkel,
  Bethard, and McClosky}]{manning-EtAl:2014:P14-5}
Christopher~D. Manning, Mihai Surdeanu, John Bauer, Jenny Finkel, Steven~J.
  Bethard, and David McClosky. 2014{\natexlab{b}}.
\newblock \href{https://doi.org/10.3115/v1/P14-5010}{The {Stanford} {CoreNLP}
  natural language processing toolkit}.
\newblock In {\em Proceedings of 52nd Annual Meeting of the Association for
  Computational Linguistics: System Demonstrations\/}. Association for
  Computational Linguistics, Baltimore, Maryland, pages 55--60.
\newblock
  \href{https://doi.org/10.3115/v1/P14-5010}{https://doi.org/10.3115/v1/P14-5010}.

\bibitem[{Meyer et~al.(2012)Meyer, Popescu-Belis, Hajlaoui, and
  Gesmundo}]{AMTA-2012-Meyer}
Thomas Meyer, Andrei Popescu-Belis, Najeh Hajlaoui, and Andrea Gesmundo. 2012.
\newblock \href{http://www.mt-archive.info/AMTA-2012-Meyer.pdf}{{Machine
  Translation of Labeled Discourse Connectives}}.
\newblock In {\em {Proceedings of the Tenth Conference of the Association for
  Machine Translation in the Americas (AMTA)}\/}.
\newblock
  \href{http://www.mt-archive.info/AMTA-2012-Meyer.pdf}{http://www.mt-archive.info/AMTA-2012-Meyer.pdf}.

\bibitem[{Miculicich~Werlen and
  Popescu-Belis(2017)}]{miculicichwerlen-popescubelis:2017:DiscoMT}
Lesly Miculicich~Werlen and Andrei Popescu-Belis. 2017.
\newblock \href{https://doi.org/10.18653/v1/W17-4802}{Validation of an
  automatic metric for the accuracy of pronoun translation (apt)}.
\newblock In {\em Proceedings of the Third Workshop on Discourse in Machine
  Translation\/}. Association for Computational Linguistics, Copenhagen,
  Denmark, pages 17--25.
\newblock
  \href{https://doi.org/10.18653/v1/W17-4802}{https://doi.org/10.18653/v1/W17-4802}.

\bibitem[{Mitkov(1999)}]{10.2307/40006919}
Ruslan Mitkov. 1999.
\newblock \href{http://www.jstor.org/stable/40006919}{Introduction: Special
  issue on anaphora resolution in machine translation and multilingual nlp}.
\newblock {\em Machine Translation\/} 14(3/4):159--161.
\newblock
  \href{http://www.jstor.org/stable/40006919}{http://www.jstor.org/stable/40006919}.

\bibitem[{Riezler and Maxwell(2005)}]{riezler-maxwell:2005:ACL2005}
Stefan Riezler and John~T. Maxwell. 2005.
\newblock \href{https://www.aclweb.org/anthology/W05-0908}{On some pitfalls in
  automatic evaluation and significance testing for mt}.
\newblock In {\em Proceedings of the ACL Workshop on Intrinsic and Extrinsic
  Evaluation Measures for Machine Translation and/or Summarization\/}.
  Association for Computational Linguistics, Ann Arbor, Michigan, pages 57--64.
\newblock
  \href{https://www.aclweb.org/anthology/W05-0908}{https://www.aclweb.org/anthology/W05-0908}.

\bibitem[{Sennrich et~al.(2016)Sennrich, Haddow, and Birch}]{sennrich-bpe}
Rico Sennrich, Barry Haddow, and Alexandra Birch. 2016.
\newblock \href{https://doi.org/10.18653/v1/P16-1162}{Neural machine
  translation of rare words with subword units}.
\newblock In {\em {Proceedings of the 54th Annual Meeting of the Association
  for Computational Linguistics (Volume 1: Long Papers)}\/}. Association for
  Computational Linguistics, Berlin, Germany, pages 1715--1725.
\newblock
  \href{https://doi.org/10.18653/v1/P16-1162}{https://doi.org/10.18653/v1/P16-1162}.

\bibitem[{Su et~al.(2012)Su, Wu, Wang, Chen, Shi, Dong, and
  Liu}]{su-EtAl:2012:ACL2012}
Jinsong Su, Hua Wu, Haifeng Wang, Yidong Chen, Xiaodong Shi, Huailin Dong, and
  Qun Liu. 2012.
\newblock \href{http://www.aclweb.org/anthology/P12-1048}{Translation model
  adaptation for statistical machine translation with monolingual topic
  information}.
\newblock In {\em Proceedings of the 50th Annual Meeting of the Association for
  Computational Linguistics (Volume 1: Long Papers)\/}. Association for
  Computational Linguistics, Jeju Island, Korea, pages 459--468.
\newblock
  \href{http://www.aclweb.org/anthology/P12-1048}{http://www.aclweb.org/anthology/P12-1048}.

\bibitem[{Tiedemann(2010)}]{tiedemann10}
J{\"o}rg Tiedemann. 2010.
\newblock \href{http://www.aclweb.org/anthology/W10-2602}{{Context Adaptation
  in Statistical Machine Translation Using Models with Exponentially Decaying
  Cache}}.
\newblock In {\em {Proceedings of the 2010 Workshop on Domain Adaptation for
  Natural Language Processing}\/}. Association for Computational Linguistics,
  Uppsala, Sweden, pages 8--15.
\newblock
  \href{http://www.aclweb.org/anthology/W10-2602}{http://www.aclweb.org/anthology/W10-2602}.

\bibitem[{Tiedemann and Scherrer(2017)}]{tiedemann_neural_2017}
J{\"o}rg Tiedemann and Yves Scherrer. 2017.
\newblock \href{https://doi.org/10.18653/v1/W17-4811}{{Neural {Machine}
  {Translation} with {Extended} {Context}}}.
\newblock In {\em {Proceedings of the Third Workshop on Discourse in Machine
  Translation}\/}. Association for Computational Linguistics, Copenhagen,
  Denmark, {{DISCOMT}'17}, pages 82--92.
\newblock
  \href{https://doi.org/10.18653/v1/W17-4811}{https://doi.org/10.18653/v1/W17-4811}.

\bibitem[{Vaswani et~al.(2017)Vaswani, Shazeer, Parmar, Uszkoreit, Jones,
  Gomez, Kaiser, and Polosukhin}]{attention-is-all-you-need}
Ashish Vaswani, Noam Shazeer, Niki Parmar, Jakob Uszkoreit, Llion Jones,
  Aidan~N Gomez, Lukasz Kaiser, and Illia Polosukhin. 2017.
\newblock
  \href{http://papers.nips.cc/paper/7181-attention-is-all-you-need.pdf}{Attention
  is all you need}.
\newblock In {\em NIPS\/}. Los Angeles.
\newblock
  \href{http://papers.nips.cc/paper/7181-attention-is-all-you-need.pdf}{http://papers.nips.cc/paper/7181-attention-is-all-you-need.pdf}.

\bibitem[{Wang et~al.(2017)Wang, Tu, Way, and {Qun Liu}}]{wang_exploiting_2017}
Longyue Wang, Zhaopeng Tu, Andy Way, and {Qun Liu}. 2017.
\newblock \href{https://doi.org/10.18653/v1/D17-1301}{{Exploiting
  {Cross}-{Sentence} {Context} for {Neural} {Machine} {Translation}}}.
\newblock In {\em {Proceedings of the 2017 Conference on Empirical Methods in
  Natural Language Processing}\/}. Association for Computational Linguistics,
  Denmark, Copenhagen, {{EMNLP}'17}, pages 2816--2821.
\newblock
  \href{https://doi.org/10.18653/v1/D17-1301}{https://doi.org/10.18653/v1/D17-1301}.

\bibitem[{Wiseman et~al.(2016)Wiseman, Rush, and Shieber}]{wiseman2016global}
Sam Wiseman, Alexander~M Rush, and Stuart~M Shieber. 2016.
\newblock \href{https://doi.org/10.18653/v1/N16-1114}{Learning global features
  for coreference resolution}.
\newblock In {\em Proceedings of the 2016 Conference of the North American
  Chapter of the Association for Computational Linguistics: Human Language
  Technologies\/}. Association for Computational Linguistics, San Diego,
  California, pages 994--1004.
\newblock
  \href{https://doi.org/10.18653/v1/N16-1114}{https://doi.org/10.18653/v1/N16-1114}.

\end{thebibliography}
\bibliographystyle{acl_natbib}

\newpage 

\appendix

\appendix


\section{Experimental setup}

\subsection{Data preprocessing}
We use the publicly available OpenSubtitles2018 corpus~\cite{open_subtitles_2016} for English and Russian.\footnote{http://opus.nlpl.eu/OpenSubtitles2018.php} 
We pick sentence pairs with an overlap of at least $0.9$ to reduce noise in the data. For context, we take the previous sentence if its timestamp differs from the current one by no more than 7 seconds. 

We use the tokenization provided by the corpus.

Sentences were encoded using byte-pair encoding~\cite{sennrich-bpe}, with source and target vocabularies of about 32000 tokens.
Translation pairs were batched together by approximate sequence length. Each training batch contained a set of translation pairs containing approximately 5000 source tokens.

\subsection{Model parameters}

We follow the setup of Transformer base model~\cite{attention-is-all-you-need}. More precisely, the number of layers in the encoder and decoder is $N=6$. We employ $h = 8$ parallel attention layers, or heads. The dimensionality of input and output is $d_{model} = 512$, and the inner-layer of a feed-forward networks has dimensionality $d_{ff}=2048$.

We use regularization as described in~\cite{attention-is-all-you-need}.

\subsection{Optimizer}
The optimizer we use is the same as in~\cite{attention-is-all-you-need}.
We use the Adam optimizer~\cite{adam-optimizer} with $\beta_1 = 0{.}9$, $\beta_2 = 0{.}98$ and $\varepsilon = 10^{−9}$. We vary the learning rate over the course of training, according to the formula:
\begin{multline*}
l_{rate}=d^{−0{.}5}_{model}\cdot \min(step\_num^{−0.5},\\ step\_num\cdot warmup\_steps^{−1.5}) 
\end{multline*}

We use $warmup\_steps = 4000$.

\end{document}